\documentclass{article}

\usepackage{url}
\usepackage{color}
\usepackage[round]{natbib}

\usepackage{authblk}

\usepackage{graphicx}

\usepackage{amsmath, amsthm, amssymb,cite}

\begin{document}

\title{A Historical Context for Data Streams}

\author[1]{Indr\.e \v{Z}liobait\.e}
\author[2]{Jesse Read}
\affil[1]{University of Helsinki, Finland}
\affil[2]{Ecole Polytechnique, France}
\affil[]{\small{indre.zliobaite@helsinki.fi;jesse.read@polytechinque.edu}}

\maketitle

\begin{abstract}
Machine learning from data streams is an active and growing research area. 
Research on learning from streaming data typically makes strict assumptions linked to computational resource constraints, including requirements for stream mining algorithms to inspect each instance not more than once and be ready to give a prediction at any time. 
Here we review the historical context of data streams research placing the common assumptions used in machine learning over data streams in their historical context. 

\end{abstract}

\section{Introduction}

Machine learning from streaming data aims at extracting knowledge from a continuous potentially neverending source of data, and constructing a model that makes predictions. 
Examples of data streams include environmental sensing, transactional, mobility or web traffic data. Learning from data streams has been an active research area over the last decades \citep{Gama_book,MOA_book} with its fundamental concepts reaching back at least to 1950s. 

This note overviews the historical context of data streams research in computer science, in particular the foundations of handling streaming data before research focus turned to machine learning. 
We cover literature up until early 2000s considering that newer work does not yet fall under history, but rather is part of active ongoing research efforts, which have been extensively surveyed elsewhere \citep{Read23,Aguiar23,Gomes19,Kojalo19,Cardellini22}.

With this historical perspective we are particularly interested in examining the origins of several assumptions typical of current research in supervised learning from data streams, that include requirements for stream mining algorithms to inspect each instance not more than once and be ready to give a prediction at any time. 

\section{Roads to Multi-Tasking\\1950s-1960s}

Algorithms existed long before mechanical calculation \citep{Daston_book}. Instructing computers to execute algorithms needed programming languages. 


Until late 1950s computer makers used custom programming languages to instruct computers what to do. Starting commercial availability of computers brought the need for universal ways of instructing them. 
Grace Hopper, a computer scientist and United States Navy rear admiral, pioneered the concept of machine-independent programming languages\footnote{\url{https://americanhistory.si.edu/cobol/proposing-cobol}}. This gave rise to COBOL\citep{COBOL}), a COmmon, Business-Oriented Language that was a high-level English-like programming language, that could universally run on more than one manufacturer's computer. COBOL dominated much of business and government data processing for decades and is still behind in some applications today. 


Many computer scientists disliked COBOL for its monolithic structures and verbosity and lack of speed as compared to non-syntax driven compilers. 
Edsger W. Dijkstra, a legendary Dutch computer scientist and software engineer, is quoted having said that 
``the use of COBOL cripples the mind; its teaching should, therefore, be regarded as a criminal offense.'' 

In response to criticisms and to show that COBOL compilers do not have to be complicated and slow a concept of coroutines was proposed \citep{Conway63}, where high-speed and one-pass were emphasized. The idea was to segment the organization of a program into autonomous modules, coroutines. The modules would communicate only in the form of discrete items of information, and information between modules would flow only one way. ``Thus, coroutines are subroutines all at the same level, each acting as if it were the master program when in fact there is no master program. [\ldots] The coroutine notion can greatly simplify the conception of a program when its modules do not communicate with each other synchronously" \citep{Conway63}.

Coroutines are cooperatively multitasked, which means that there is never a context switch from a running unfinished process to another process. 
An alternative is preemptive multitasking, which has an interruption mechanism that can suspend the currently executing process and invokes a scheduler to determine which process should execute next. 
The advantage of cooperative multitasking is that there is no need for support from the operating system, there is no need to guard critical sections, which is particularly important in hard real-time contexts. Execution is never unexpectedly interrupted by the process scheduler, various functions inside the application do not need to be reentrant \citep{Bartel88}. 

The price to pay is a fragility of the system. A cooperatively multitasked system relies on each process regularly giving up time to other processes, in such circumstances a poorly designed routine can take excessive time or cause the whole system to hang. This must be the primary source of computational constraints and single-pass requirements quoted in machine learning research over data streams today. 

In parallel to the developments of business-oriented procedural programming languages in the 1960s, \textit{dataflow programming} arose \citep{kelly1961block} to emphasize functional programming particularly suited for numerical processing. Until 1960s data was an integral part of computer programs generated by the computer, rather than handled as a dataset from a separate source. Dataflow programming started seeing a program as a directed graph where data flows between operations \citep{Sutherland66}; in that sense the beginnings of dataflow architectures can be seen as early stream processing. 

\section{Data Streams in Computing and Processing\\ 1970s-1990s}

The keyword \emph{data streams} emerged in the 1970s in the context of modular, structural programming \citep{Burge75,Morrison71,Morrison78}. 
Streams were introduced as functional analogues to coroutines. 
``A stream is [\ldots] a particular method of representing a list in which the creation of each list element is delayed until it is actually needed." \citep{Burge75}. 

Streaming algorithms were studied by \citep{MUNRO1980315} as early as 1978, motivated by the context where ``a file stored on a read-only tape and the internal storage is rather limited, several passes of the input tape may be required''; but even in this case made use of internal storage. As well as \citep{flajolet1985approximate} and others. 

In the streams context, the program would be written as if it were a multipass program, but when executing, the parts of the separate passes would be interleaved. An example of that is when a function produces a list in a natural order and another function processes the list items in the same order, it is not necessary to produce the whole list before applying the second function to it. 
The concept of stream is thus primarily related to operations on demand without prior contingency. This implies the absence of a buffer to memorize and store interim information. 

In the context of computing the keyword `streams' appeared as data stream linkage mechanism (DSLM) \citep{Morrison78}, where the idea was to link program modules from a network through which data passes. In DSLM framework two or more communicating ports were connected by a buffer called a queue. The key was that none of the processes needed the identity of the preceding or succeeding process. This made the system very modular and portable, in principle the modules should be easily reconfigurable. The modularity aspect is pretty much lost in machine learning over data streams today.

Meanwhile during the 1980s, data stream processing continued to be developed in the context of data flow programming. 
A general purpose single assignment programming language SISAL was developed in that decade \citep{SISAL} that converted a textual human readable dataflow language into a graph format.

Also around this time we have the beginnings of a split to what could be considered a precursor to algorithms designed specifically for \emph{learning from} (machine learning) versus \emph{processing} and otherwise managing data streams, which involved classical computer algorithms like selection and sorting; typically done in a single pass, with limited storage, although several passes were permitted if resources allowed.

Of course at their core, data-stream learning is a form of data-stream processing; and this will never be a clean split. We can highlight that Rosenblatt's perceptron \citep{rosenblatt1957perceptron} is essentially a data-stream processing algorithm; at the time memory was indeed extremely scarce. However, as computing resources became increasingly ubiquitous, learning began to grown as a separate focus than simply processing. 

Typical stream algorithms included ways to calculate a running mean or a similar counting-based measurement. 
But for more complex measurements and faster arriving larger streams, exact computation became impossible, and the focus shifted on data summaries (often called sketches) \citep{flajolet1985approximate}, optimized for approximately answering associated types of queries. 

Further formalization and popularization of streaming algorithms came from a 1996 paper by \citep{alon1996space}. By this time, computing resources had become orders of magnitude more powerful and plentiful than previous decades, and data transfer between computers became more pressing than data transfer among components of the same computer.

\section{Very Large Databases and Rapid Data\\ 1960s-1990s-2000s}

Similarly to algorithms, the concept of databases existed well before there were computers in a form of card catalogues or alike. Databases, electronic or not, would have four major components: 
data, a data structure, a medium in which to exist, and routines (algorithms) to sort and access that data \citep{Sepkoski17}. Charles Bachman, an American computer scientist, designed the first computer database system in 1960s. In 1970s a relational model of handling data was introduced \citep{Codd70}. The premier conference on Very Large Databases\footnote{\url{https://vldb.org/}} started soon after. And while notion of what makes data very large certainly changed over the years, the community remains more active than ever and will celebrate its 50 years anniversary next year. 

Rapid arrival of data came prominently into research in late 1990s, coinciding with rapid expansion of the internet (and the dot-com bubble, towards the end of the decade), and massive digitization of data. Data could be passed from one computer to another with unprecedented ease. Suddenly there was an exponential increase in the amount of data that could be processed, even greater than CPU (Moore's Law) or hard-disk space could keep up with.

From the data base community the term `data stream management system (DSMS)' arrived, where things could be expressed as a continuous query \citep{chen2000niagaracq}. 



This led to the term `data streams' departing from its attachment to software development; and it became primarily associated with arrival and management of data in databases \citep{Babcock02}, as well as extracting knowledge from data. \citet{Babcock02} and references therein mention application in searching and evaluating, intrusion detection,  tracking, analysis, and monitoring; frequently referring to the web and sensor networks as the main sources of such tasks and data.

Rapid arrival and potentially unbounded volumes of data were contrasted to the traditional database management systems and the term massive data streams came into popularity \citep{Domingos03,Muthukrishnan05}. 
In the same publication \citet{Domingos03} formulated requirements for systems mining high-volume open-ended data streams:
\begin{itemize}
\item A system must require small constant time per record, otherwise it will inevitably fall behind the data, sooner or later. 
\item It must use only a fixed amount of main memory, irrespective of the total number of records it has seen. 
\item It must be able to build a model using at most one scan of the data, since it may not have time to revisit old records, and the data may not even all be available in secondary storage at a future point in time. 
\item It must make a usable model available at any point in time, as opposed to only when it is done processing the data, since it may never be done processing. 
\item Ideally, it should produce a model that is equivalent (or nearly identical) to the one that would be obtained by the corresponding ordinary database mining algorithm, operating without the above constraints. 
\item When the data-generating phenomenon is changing over time (i.e., when concept drift is present), the model at any time should be up-to-date, but also include all information from the past that has not become outdated.
\end{itemize}

In that manifesto \citet{Domingos03} postulated that ``even if we have (conceptually) infinite data available,
it may be the case that we do not need all of it to obtain the best possible model of
the type being mined." They asked ``assuming the data-generating process is stationary, is there
some point at which we can “turn off” the stream and know that we will not lose predictive performance by ignoring further data?". Here not losing predictive accuracy, is meant in the practical sense, theoretically, achieving asymptotically optimal error requires infinite amount of data and there is an incremental improvement in generalization with ever more training samples \citep{Vapnik_book}.

\citet{Domingos03} explicitly discussed time changing data streams: ``if the data-generating process is not stationary, how do we make the trade-off between being up-to-date and not losing past information that is still relevant?''

While the phenomenon of concept drift has by now been studied for secveral decades in the supervised machine learning context  \citep{Widmer96}, the focus in concept drift research then was on learning accurate representations from limited amount of information. \citet{Hulten01} linked concept drift to potentially very large data.

\section{Lead up to Present Day Contexts\\
1990s--2000s--2020s}


In the early 2000s the term `internet of things' (IoT) came into use, primarily focusing on devices with sensors typically connected into communication networks.  
IoT, and rapid development of the internet and networks in general continued motivate data streams research in academic and industrial contexts.


Data streams research was further catalyzed by the term "Big Data" and the so-called 3 Vs (volume, velocity and variety; later increased to 4 (veracity) and 5 (value) Vs) \citep{laney20013d} that entered into marketing use around 2001-2002. 
The research area that is known today as \textit{data stream learning} was driven along at hectic speeds by exploasion of attention towards Big Data and developments in machine learning followed this unprecedented attention to data \citep{WEKA}. 

In this modern context, rather than streams arising out of software development, software development arose to learn from streams \citep{bifet2010moa}. Learning in this context was, at this time (by 2010s) almost entirely under the umbrella of what we now see as the traditional machine learning and its variations, including decision trees, dense shallow neural networks (often with no hidden layer), nearest-neighbours and Bayesian learninig approaches.

By the 2020s deep learning has matured and also intersected with the omnipresence of streams, but under different perspective, such as continual "life-long" learning, or deep reinforcement learning.

\section{Concluding Remarks}


The concept of data streams transitioned over the years from multitasking and modularity in software engineering to effective knowledge discovery from very large datasets. 

The requirement for a single pass in knowledge discovery from large databases  was formulated as a rational desiderata rather than a boundary condition in a form of hard constraint. 

While the history of data streams research goes back to the earliest days of computing, the setting remains topical and timely at present day, especially in the context of advanced and ever more accessible machine learning. 

Early authors recognized that in practice, streams are finite, and knowledge discovery systems may not need to learn continuously and infinitely; but can persist with timely updates in the case of concept drift. 

We support this view; there is now sufficient data available for algorithms to learn an incredible wealth of knowledge, without the inherently risky task of continuous updating and adaptation.

That said, many challenges and open problems remain. We hope for many new research insights in the years to come.

\bibliographystyle{named}
\bibliography{note_historical_streams}

\end{document}